\documentclass[10pt,conference,a4paper]{IEEEtran}
\usepackage{cite}
\usepackage[pdftex]{graphicx}
\usepackage{amsmath}
\usepackage{algorithm, algorithmic}
\usepackage{array}

\usepackage{tabulary}
\usepackage{multirow} 

\begin{document}
\title{Multiple Instance Learning Convolutional Neural Networks for Object Recognition}


\author{\IEEEauthorblockN{Miao Sun\IEEEauthorrefmark{1},
Tony X. Han\IEEEauthorrefmark{1},
Ming-Chang Liu\IEEEauthorrefmark{2} and 
Ahmad Khodayari-Rostamabad\IEEEauthorrefmark{2} 
\IEEEauthorblockA{\IEEEauthorrefmark{1}Electrical and Computer Engineering\\
University of Missouri,
Columbia, Missouri 65211\\ Email: msqz6@mail.missouri.edu,  hantx@missouri.edu}
\IEEEauthorblockA{\IEEEauthorrefmark{2}Sony Electronics Inc, San Jose, California, 95112\\
Email: Ming-Chang.Liu, Ahmad.Khodayari@am.sony.com}
}}

\maketitle

\begin{abstract}
Convolutional Neural Networks (CNN) have demonstrated its successful applications in computer vision, speech recognition, and natural language processing. 
For object recognition, CNNs might be limited by its strict label requirement and an implicit assumption that images are supposed to be target-object-dominated for optimal solutions. 
However, the labeling procedure, necessitating laying out the locations of target objects, is very tedious, making high-quality large-scale dataset prohibitively expensive.
Data augmentation schemes are widely used when deep networks suffer the insufficient training data problem.
All the images produced through data augmentation share the same label, which may be problematic since not all data augmentation methods are label-preserving.
In this paper, we propose a weakly supervised CNN framework named \textit{Multiple Instance Learning Convolutional Neural Networks} (MILCNN) to solve this problem.
We apply MILCNN framework to object recognition and report state-of-the-art performance on three benchmark datasets: CIFAR10, CIFAR100 and ILSVRC2015 classification dataset.
\end{abstract}


\IEEEpeerreviewmaketitle

\section{Introduction}
\label{sec:intro}


Deep learning algorithms\cite{DlNature} are revolutionizing various tasks in artificial intelligence including natural language processing\cite{NLPS, QASE, VNMT, SSNN} , speech recognition\cite{NNLM, DNNSR, DcnnSR}, and computer vision\cite{Alex, HieSceLab, JoinPose}. 
The successes of deep learning algorithms are the result of its excellent capability to discover intricate structures in high-dimensional data with little manual engineering. 
The breakthroughs in ImageNet challenge\cite{ILSVRC12} has demonstrated that powerful feature representations can be learned from data automatically, outdating traditional approaches based on hand-designed features. 

The most successful algorithm in deep learning algorithms for image recognition is Convolutional Neural Networks (CNN)\cite{Alex, GoogleNet, Frcnn, ResNet}. 
The architecture of typical CNNs\cite{LecunGrad} is a stack of convolutional, non-linear, pooling and fully-connected layers, followed by a loss function layer. It is designed to take advantages of local connections, shared weights, pooling and the use of many layers to learn high-level representations of natural images, and it has demonstrated significant improvement over various benchmark object recognition datasets \cite{ILSVRC12, Cifar, PASCAL07}. 
However, deep CNNs necessitate large well-labeled training data to achieve these superior results, whereas the labeling work is very tedious and expensive by hand. Limited amount or inferior quality of training data will lead to suboptimal models.
Currently, ImageNet, the largest labeled high-resolution image database available publicly,  has around 14 million images and 22 thousand synsets. ILSVRC2015 is a subset of ImageNet with roughly 1,300 images in each of 1000 categories. In all, there are about 1.3 million training images, 50,000 validation images, and 150,000 testing images. 
For deep CNNs with million-level parameters, ILSVRC2015 is the most widely used dataset for training deep convolutional neural networks.
Krizhevsky et. al~\cite{Alex} used a simple data augmentation scheme of generating image translations and horizontal reflections to increase the size of the 1.3 million training set by 2048 times. Without the scheme, deep networks would suffer substantial overfitting. Thus, this straightforward and powerful scheme is considered to be default for recent deep CNNs~\cite{GoogleNet, Frcnn, ResNet}. \cite{HCP, lcnn} showed the promising performance could be achieved by high-quality bounding box proposals instead of randomly cropping images. 

\begin{figure}[t]
\centering
\includegraphics[width=2.5in]{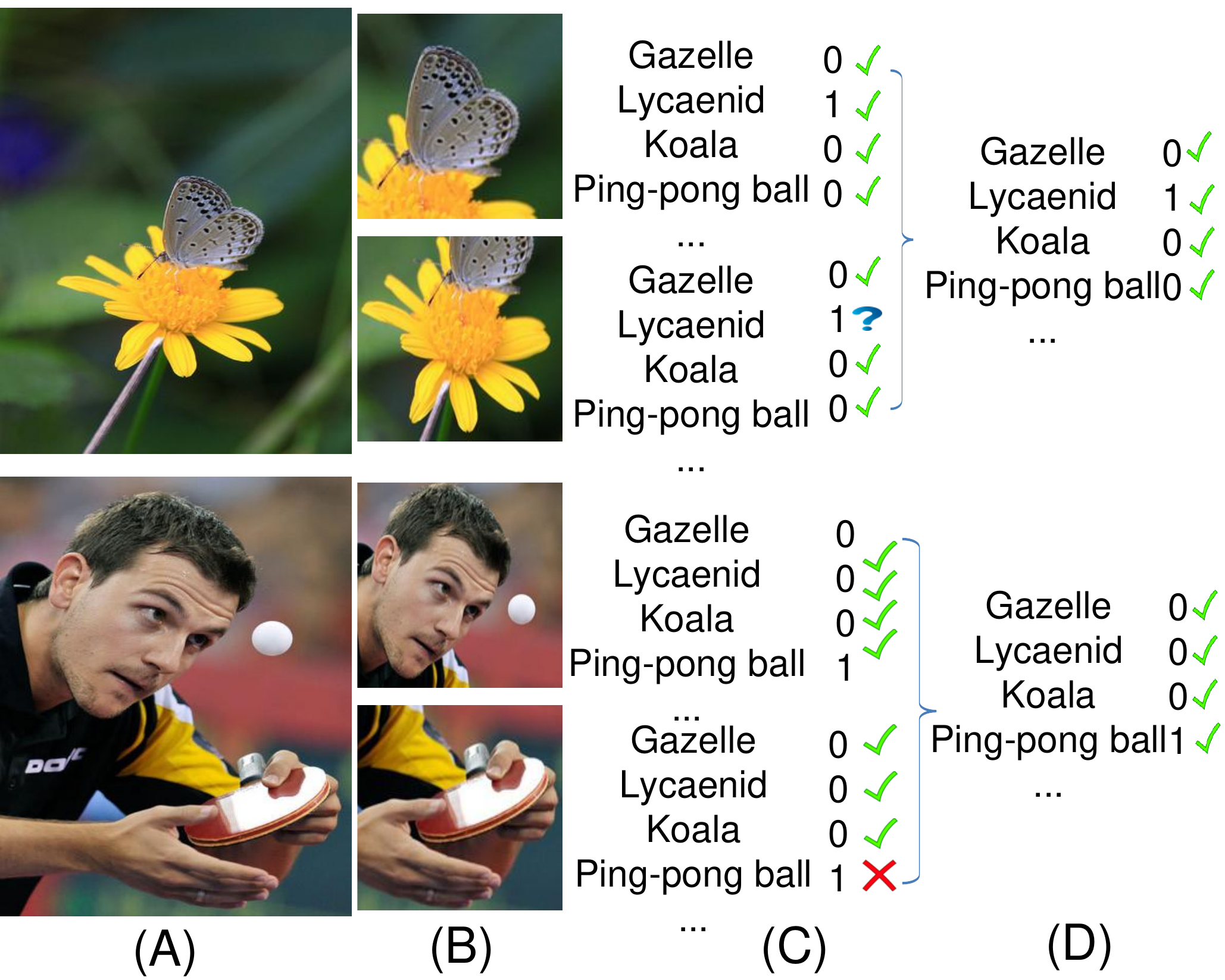}
\caption{{\bf Data augmentation and labels:} (A) Original images from the ILSVRC2015 training set. (B) Image regions created for deep CNNs by data augmentation. (C) Labels assigned to created image areas. Check marks represent high-quality label. The question mark is a low quality label. Cross marks mean incorrect labels. (D) Image-level labels. All created regions share the same image-level label, which is intrinsically associated with multiple instance learning.}
\label{fig:intro}
\end{figure}

In practice, not all data augmentation methods are label-preserving, especially when target objects fail to dominate the whole images. As a supervised learning method, CNNs take (data, label) pairs as in Fig~\ref{fig:intro} (B-C). The inferior quality labels make it difficult for optimization and lead to suboptimal CNN models. 
To reduce the effect of noisy training pairs, we model deep CNNs in a weakly supervised learning framework. Instead of assigning labels to all generated images (as in Fig~\ref{fig:intro}(C)), we treat the generated images as a bag and the original label as bag-level label (as in Fig~\ref{fig:intro}(D)).
This phenomenon is intuitively the problem of Multiple Instance Learning (MIL). For binary MIL, a bag is labeled positive if the bag contains at least one positive instance, and it is labeled negative if all its instances are negative. Therefore, incorporating MIL into deep learning algorithm would fully utilize the potential of the training set and achieve better performance.

Recently, He et al.\cite{ResNet} built a 152-layer deep network for high-resolution images\cite{ILSVRC12} and over 1000-layer deep network for low-resolution images~\cite{Cifar} by taking advantages of carefully designed initialization and residual learning techniques. The significant increase of depth of layers results in outperforming previous CNNs in various benchmark datasets\cite{ILSVRC12, PASCAL07, COCO}. 

In this paper, we propose a weakly supervised CNN framework named \textit{Multiple Instance Learning Convolutional Neural Networks} (MILCNN) to reduce the effect of noisy labels. The proposed algorithms are validated by extensive experiments and have shown state-of-the-art performance on various object recognition benchmarks.  

\section{MILCNN for Image Classification}

In this section, we will first briefly introduce mathematical formulations of CNNs including how to predict labels and update parameters. 
Then, we will give detailed equations about how to incorporate multiple instance learning into CNN. 
The whole procedure is summarized in Algorithm~\ref{algm:MILCNN}. 

\subsection{Traditional CNN Formulation}
Convolutional Neural Networks (CNN) are a special kind of neural networks, which consist a stack of convolutional layers, pooling layers, fully-connected layers and loss function layers. Optimization of CNNs is a supervised learning process to minimize loss function:

\begin{equation} \label {eq:allloss}
   L = \sum_{i=1}^{N} L_i =  \sum_{i=1}^{N} f_{loss}(\mathbf{y_i}, F(\mathbf{W}, \mathbf{x_i}))
\end{equation}

$L$ is the total cost of $N$ training examples. $\mathbf{y_i} = \{0, 1\}^{1 \times C}$ is the label matrix, where $C$ is the number of categories. $\mathbf{W}$ represents the collection of adjustable parameters in the structure. $\mathbf{x_i} \in \Re^{1 \times D}$ is the input pattern matrix, where $D$ is the dimension of each input pattern. $F(\mathbf{W}, \mathbf{x_i}) \in \Re^{1 \times C} $ can be interpreted as the category label of input patterns or probabilities associated with each category. $F$ indicates a set of functions in CNN such as convolutional functions, pooling functions and fully-connected functions. 
For notational convenience,  we will drop the example index and use $\mathbf{x}$ to denote the chosen input data and $\mathbf{y}$ the corresponding label. Then the Eq (\ref{eq:allloss}) will be 

\begin{equation} \label {eq:oneloss}
   L = f_{loss}(\mathbf{y}, F(\mathbf{W}, \mathbf{x}))
\end{equation}

The most widely used loss function in convolutional neural networks is softmax with cross-entropy loss function.
Let CNN output $\mathbf{h}=F(\mathbf{W}, \mathbf{x})$, $\mathbf{h} = \{h_1, h_2, ..., h_C\}$, the predicted label is max value of $\mathbf{h}$

\begin{equation} \label {eq:label}
   \hat y = {argmax}_{i=1}^{C} (h_i)   
\end{equation}

and the cross-entropy loss is

\begin{equation} \label {eq:softmaxentropy}
   f_{loss}= - \sum_{i=1}^{C} y_i \log (p_i)
\end{equation}

where $p_i = \frac{\exp(h_i)}{\sum_{j=1}^{C} \exp(h_j)}$ for  $i=1, 2, ...C$

The gradient of softmax with cross-entropy loss w.r.t $\mathbf{h}$ is 

\begin{equation} \label {eq:softmaxentropy_grad}
   \frac{\partial{f_{loss}}}{\partial{h_i}}= -y_i + p_i \sum_{j=1}^{C} y_j
\end{equation}

According to chain rules, we will have 
\begin{equation} \label {eq:weights_grad}
   \frac{\partial{f_{loss}}}{\partial{\mathbf{W}}}= \frac{\partial{f_{loss}}}{\partial{\mathbf{h}}}  \frac{\partial{\mathbf{h}}}{\partial{\mathbf{W}}}
\end{equation}

The parameters of the CNN is updated as
\begin{equation} \label {eq:weights_update}
 \mathbf{W}_{new} = \mathbf{W}_{old} - \lambda \frac{\partial{f_{loss}}}{\partial{\mathbf{W}}} 
\end{equation}

In the simplest case, the learning rates $\lambda$ is a scalar constant. 

\subsubsection{MILCNN Formulation}
\label{sec:milcnn}

For multiple instance learning, training instances are not singletons. Instead, they come in ``bags", where all the examples in a bag share the same label. For object recognition, regions in each image are considered as a bag as in Fig~\ref{fig:intro}(B). 

Let $B=\{\mathbf{x}^1,\mathbf{x}^2 ... \mathbf{x}^m\}$, $\mathbf{x}^m$ is $m_{th}$ region in the image. The loss function w.r.t to the bag $B$ is

\begin{equation} \label {eq:bagloss}
   f_{loss}= - \sum_{i=1}^{C} y_i \log (p(c_i=1|B))
\end{equation}

where $p(c_i=1|B)$ represents the probability that the bag is classified into $i_{th}$ category. According to concept of multiple instance learning, $B$ is a negative bag for the $i_{th}$ category if all the instances in the bag are negative:

\begin{equation} \label{eq:negbag}
  p(c_i=0|B) = \Pi_{j=1}^{m}(1-p(c_i=1|\mathbf{x}^j))
\end{equation}

$p(c_i=1|\mathbf{x}^j)$ is the probability of $j_{th}$ region to be considered as $i_{th}$ category, and

\begin{equation} \label{eq:px}
  p(c_i=1|\mathbf{x}^j) = 1 - \exp (-\lambda h_i^j)
\end{equation}

where $h_i^j$ is the $i_{th}$ output of CNN model before loss layer for $j_{th}$ region , and $h_i^j \in [0, \infty)$ (if there is a ReLU layer before the loss function layer), $\lambda$ is constant positive value. The reason why we define the probability as Eq (\ref{eq:px}) is to simplify calculation of gradients below. 

Optimization of Eq (\ref{eq:bagloss}) is equivalent to minimization of the following function 
\begin{equation} \label {eq:negbagloss}
   f_{loss}= - \sum_{i=1}^{C} (1-y_i) \log (1-p(c_i=1|B))
\end{equation}

Combining Eq (\ref{eq:negbagloss}) with Eq (\ref{eq:px}, \ref{eq:negbag}) will have

\begin{equation} \label {eq:negbagloss2}
\begin{split}
   f_{loss} &\quad = - \sum_{i=1}^{C} (1-y_i) \log (1-p(c_i=1|B)) \\
            &\quad = - \sum_{i=1}^{C} (1-y_i) \log (p(c_i=0|B)) \\
            &\quad = - \sum_{i=1}^{C} (1-y_i) \log \Pi_{j=1}^{m}(1-p(c_i=1|\mathbf{x}^j)) \\
            &\quad = - \sum_{i=1}^{C} (1-y_i) \sum_{j=1}^{m} \log (1-p(c_i=1|\mathbf{x}^j)) \\
            &\quad = - \sum_{i=1}^{C} (1-y_i) \sum_{j=1}^{m}\log( \exp (-\lambda h_i^j)) \\ 
            &\quad = - \sum_{i=1}^{C} (1-y_i) \sum_{j=1}^{m}(-\lambda h_i^j) 
\end{split}
\end{equation}

Then the corresponding gradients will be 
\begin{equation} \label {eq:bag_grad}
   \frac{\partial{f_{loss}}}{\partial{h_i^j}}=\lambda (1-y_i) 
\end{equation}

Given all the equations above, MILCNN can be summarized in Algorithm~\ref{algm:MILCNN}

\begin{algorithm}[htp]
  \caption{\small{MILCNN for Image Classification}}
  \label{algm:MILCNN}
  \begin{algorithmic}
    \STATE {\bfseries INPUT:} Image-label pairs ($\mathbf{x_i}$, $\mathbf{y_i}$), $i = 1 ... N$, and training epoch number $EN$.
    \STATE {\bfseries OUTPUT:} CNN parameters $\mathbf{W}$, Predicted labels $\hat {\mathbf{y}} $.

    \STATE $\mathbf{Training}$
    \FOR{$j=1$ {\bfseries to} $EN$}
      \FOR{$i=1$ {\bfseries to} $N$}
        \STATE 1. Create image bags by data augmentation and transfer given labels to vector format.
        \STATE 2. Compute loss for each bag with Eq (\ref{eq:oneloss}, \ref{eq:negbagloss2}).
        \STATE 3. Compute gradients for each bag with Eq (\ref{eq:weights_grad}, \ref{eq:bag_grad}).
        \STATE 4. Update CNN parameters with Eq (\ref{eq:weights_update}).
      \ENDFOR
    \ENDFOR
    \STATE $\mathbf{Testing}$
      \FOR{$i=1$ {\bfseries to} $N$}
        \STATE 1. Create image bags by data augmentation and transfer given labels to vector format.
        \STATE 2. Compute CNN output $\mathbf{h}$ before loss layer.
        \STATE 3. Predict label with Eq (\ref{eq:label}) or possibilities with Eq (\ref{eq:px}).

      \ENDFOR
  \end{algorithmic}
\end{algorithm}

{\bf Discussion of predicted possibilities:}
In the conventional setup, the output of softmax layer is considered as predicted possibilities. 
However, softmax layer is only suitable for single label problem, which means the input image only contains one target object. 
For example, if one image $\mathbf{x}$ has target objects, such as ``Ping-Pong ball" and ``Person", it will confuse the model because  $\max_{i=1}^{C} p_i$ is trying to reach $0.5$ instead of $1$ for the ground truth category. Therefore, we discard the softmax layer and use Eq (\ref{eq:px}) in the MILCNN framework instead. 
Eq (\ref{eq:px}) indicates that there are chances that input images can contain multiple target objects or contain no target objects at all. Compared to softmax function, the possibility is not based on one input image but related to all training images.

\section{Experiments}
In this section, we will first talk about detailed configurations of MILCNN structures, and then we will provide state-of-the-art performance on benchmark datasets: CIFAR10, CIFAR100 and ILSVRC2015 classification dataset. 

\subsection{MILCNN Structures}

\begin{figure}[h]
\centering
\includegraphics[width=0.9\linewidth]{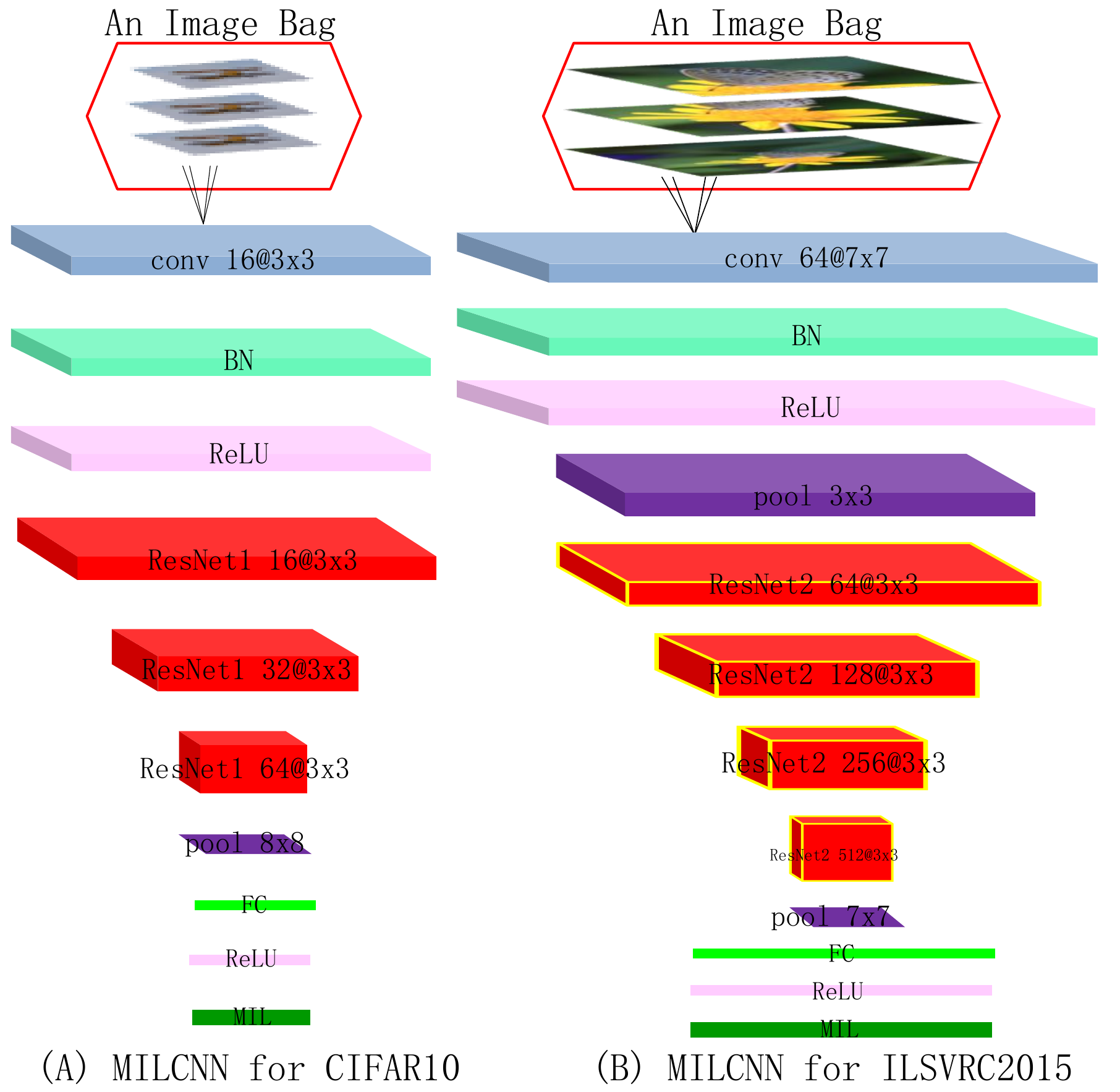}
\caption{{\bf MILCNN Structures:} Use images bags for both structures. conv $16@3 \times 3$ means there are $16$ convolutional kernels with receptive field size $3$ by $3$. Two different residual network blocks are used for two datasets separately and details in residual network blocks are illustrated in Fig~\ref{fig:resnetblock}. }
\label{fig:framework}
\end{figure}

Our MILCNN is a combination of deep residual network~\cite{ResNet} and multiple instance learning loss layer as in Fig~\ref{fig:framework}. MILCNN is a stack of layers including convolutional layers, batch normalization layers~\cite{BN}, rectified linear unit (ReLU)~\cite{ReLU}, residual network layers~\cite{ResNet}, pooling layers, fully-connected layers and multiple instance learning loss function, which is developed in Section~\ref{sec:milcnn} \footnote{The detailed residual networks are based on the work of Sam Gross and Micheal Wiber called ``Training and investigating Residual Nets".}.

Convolutional layers make use of local connections and shared weights, which is effective to avoid overfitting compared to fully-connected layers. Thus, we only use one fully-connected layer to project feature maps to the number of categories of a training set. 
Batch Normalization has been verified to be able to accelerate deep networks training via reducing covariate shift. 
ReLU is used to replace traditional nonlinearity such as sigmoid and tangent functions, which tend to slow down learning procedure due to saturating effect. 
Multiple instance learning loss layer is developed in this paper to work harmony with data augmentation to fully explore the potential of training sets. 

\begin{figure}[!t]
\centering
\includegraphics[width=2.5in]{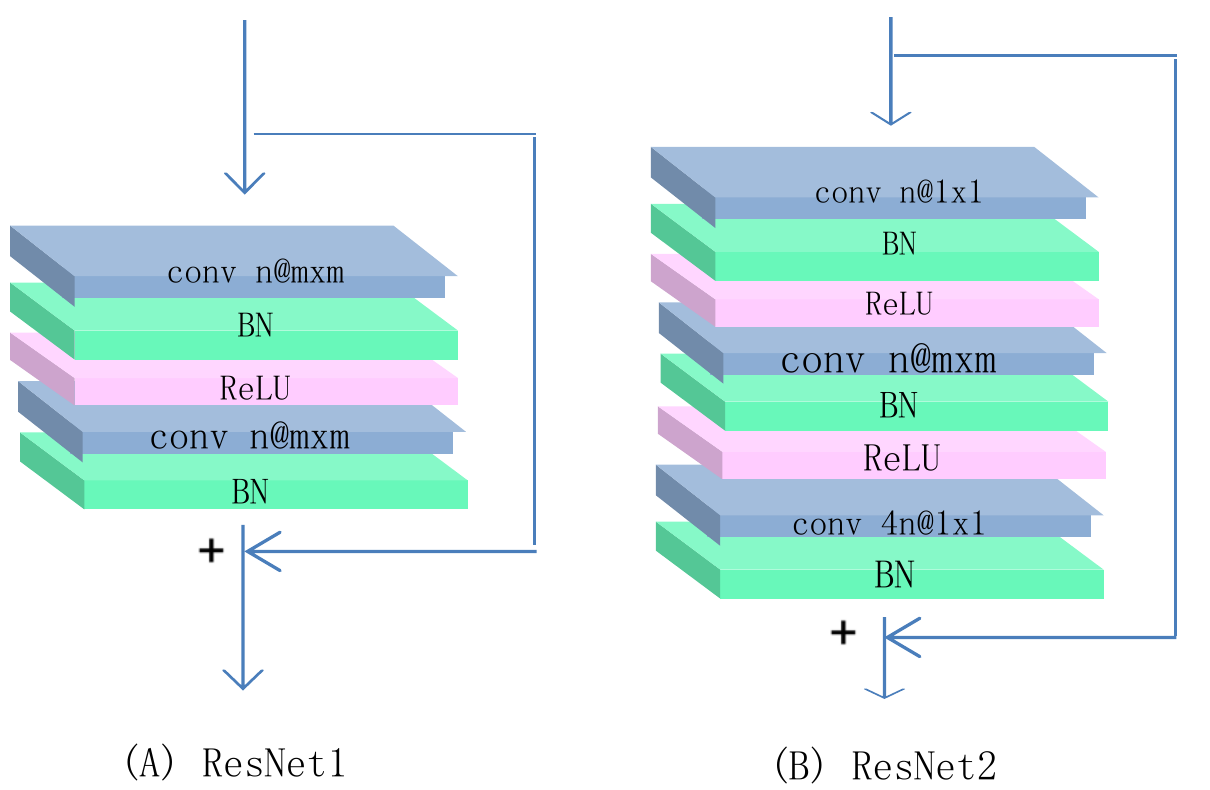}
\caption{{\bf Residual Network Blocks}: Two different residual network block types used in Fig~\ref{fig:framework}. conv $n@m \times m$ means $n$ convolutional kernel with size $m$ by $m$.}
\label{fig:resnetblock}
\end{figure}

With network depth increasing, accuracy gets saturated and then degrades rapidly, that is, degradation problem. Degradation is not caused by overfitting but related to the difficulty of optimization of substantial deep neural networks. 
This issue is recently addressed by residual networks. ~\cite{ResNet} shows it is much easier to optimize the residual mapping with reference mapping than to optimize the original, unreferenced mapping. 
Figure~\ref{fig:resnetblock} shows two implementations of residual networks using shortcut connections. Shortcut connections are those skipping one or more layers, and identity mapping is most widely used as shortcut connections. The stacked nonlinear layers in Fig~\ref{fig:resnetblock} is therefore called residual mapping. 

\subsection{CIFAR10} \label{sec:cifar10}
CIFAR10~\cite{Cifar} is a dataset of RGB images containing 32 x 32 pixels. It has 10 categories with 50,000 training images and 10,000 test images. 

Our best MILCNN structure for CIFAR10 is in Fig~\ref{fig:framework} (A) and detailed layer setup is summarized in Table~\ref{tab:milcifar}. 
Configurations of residual network blocks are shown in the brackets. In the bracket, two $3 \times 3, 16$ represents 16 convolutional kernels with receptive field size 3 by 3 for Type 1 residual network block in Fig~\ref{fig:resnetblock}(A). For each convolutional layer, we choose stride 1 and padding size 1 so that the output feature map will not change. The downsampling is only performed by first convolutional layer in ResNet2 and ResNet3 with a stride of 2, so the output size will decrease by a factor of 2. Batch normalization layer and ReLU layer do not affect sizes of feature map and thus they are not listed in this table.
We refer this configuration as 111-layer following convection of ~\cite{ResNet} by excluding batch normalization layers, ReLU layers and loss layer. If taking all these layers into consideration, the total number of layers will be $1 + 1 + 1 + 5 \times 18 \times 3 + 1 + 1 + 1 + 1 = 277$.

\newcommand{\blocka}[2]{\multirow{3}{*}{\(\left[\begin{array}{c}\text{3$\times$3, #1}\\[-.1em] \text{3$\times$3, #1} \end{array}\right]\)$\times$#2}
}
\renewcommand\arraystretch{1.1}
\setlength{\tabcolsep}{3pt}
\begin{table}[h]
\caption{CNN architecture for CIFAR10.}
\begin{center}
\resizebox{0.7\linewidth}{!}{
\begin{tabular}{c|c|c}
\hline
layer name & output size & 111-layer \\
\hline
conv1 & 32$\times$32 & 3$\times$3, 16, stride 1, padding 1\\
\hline
\multirow{3}{*}{ResNet1} &  \multirow{3}{*}{32$\times$32}  & \blocka{16}{18}  \\ 
&  & \\
&  & \\
\hline
\multirow{3}{*}{ResNet2} &  \multirow{3}{*}{16$\times$16}  & \blocka{32}{18}  \\ 
&  & \\
&  & \\
\hline
\multirow{3}{*}{ResNet3} &  \multirow{3}{*}{8$\times$8}  & \blocka{64}{18}  \\ 
&  & \\
&  & \\
\hline
pool2 & 1$\times$1  & average pool 8$\times$8  \\
\hline
fc & 10 & fc size 64$\times$10  \\
\hline
\end{tabular}
}
\end{center}
\vspace{-.5em}
\label{tab:milcifar}
\vspace{-.5em}
\end{table}
During training of this MILCNN, we use a weight decay of 0.0001 and momentum 0.9. The initial learning rate is 0.1 and reduces to 0.01, 0.001 at 80 epochs and 120 epochs. We form image bags by sampling 5 $32 \times 32$ regions from a 4-pixel padded images, that is, the image resolution is $40 \times 40$. The best performance is 5.1\% error rate and achieves superior results compared to other state-of-the-art methods as shown in Table~\ref{tab:cifar10}.

\begin{table}[h]
  \caption{Test set error rates for CIFAR-10 of various methods}
  \label{tab:cifar10}
  \vskip 0.15in
  \begin{center}
  \begin{small}
  \begin{sc}
  \begin{tabular}{p{6cm} c}
    \hline
    Method                                                     & Test Error \\
    \hline
    Conv. maxout~\cite{Maxout}   &  9.38\% \\
    DropConnect + 12 networks~\cite{Dropcon} &  9.32\% \\
    NIN ~\cite{NIN}               &  8.81\% \\
    ResNet(110-layer)~\cite{ResNet}                        & 6.43\% \\
    \hline
    \hline
    ResNet (111-layer)                  & 6.07\% \\
    MILCNN (111-layer)                  & 5.11\% \\
    
  \end{tabular}
  \end{sc}
  \end{small}
  \end{center}
  \vskip -0.1in
\end{table}

\subsection{CIFAR100}
CIFAR100~\cite{Cifar} is a dataset of RGB images containing 32 x 32 pixels. It has 100 categories with 50,000 training images and 10,000 test images. 

Our best MILCNN structure for CIFAR100 is similar to Fig~\ref{fig:framework} (A) and Table~\ref{tab:milcifar}, except that the last fully-connected layer has 100 output nodes. The training hyper-parameters and procedure are also similar to Secion~\ref{sec:cifar10}.

\begin{table}[h!]
  \caption{Test set error rates for CIFAR-100 of various methods}
  \label{tab:cifar100}
  \vskip 0.15in
  \begin{center}
  \begin{small}
  \begin{sc}
  \begin{tabular}{p{6cm} c}
    Method                                                     & Test Error \\
    \hline
    Learned Pooling~\cite{LearnedPool}                          & 43.71\% \\
    Stochastic Pooling~\cite{StoPool}                          & 42.51\% \\
    Conv. maxout ~\cite{Maxout}                      & 38.57\% \\
    Tree based priors~\cite{TreePrior}                         & 36.85\% \\
    NIN~\cite{NIN}                                   & 35.68\% \\
    \hline
    ResNet (111-layer)                                   & 28.54\% \\
    MILCNN (111-layer)                                   & 26.42\% \\
  \end{tabular}
  \end{sc}
  \end{small}
  \end{center}
  \vskip -0.1in
\end{table}

In Table~\ref{tab:cifar100}, ResNet (111-layer) has achieved the state-of-the-art performance with the help of residual network blocks. However, after incorporating our multiple instance learning loss layer, the performance has a significant improvement to 26.42\% for a single model.

\subsection{ILSVRC2015}

ILSVRC2015 is a subset of ImageNet classification dataset, and it contains 1.28 million training images and 50,000 validation images. 

Our best MILCNN structure for ILSVRC2015 is in Fig~\ref{fig:framework} (B) and detailed layer setup is summarized in Table~\ref{tab:mililsvrc}.
Compared to configurations in Fig~\ref{fig:framework} (A), we add one pooling layer after the first convolutional layer and choose the second type residual network block in Fig~\ref{fig:resnetblock}. These modifications are mainly due to high-resolution images, whose resolution is 224 by 224 (The original resolution of images in ILSVRC2015 classification dataset is around 500 by 400, we resize and crop $224 \times 224$ for our CNN architecture input).
In the brackets, convolutional kernels with receptive field size $1 \times 1$ are adopted to reduce parameters of CNN models. For shortcut connections, type2 residual network blocks do not use identity mapping but a project matrix implemented via $1 \times 1$ convolutions to match dimensions. The downsampling is only performed by middle convolutional layer in ResNet2,  ResNet3 and ResNet4 with a stride of 2, so the output feature map will decrease by a factor of 2.
We refer this configuration as 103-layer following convection of ~\cite{ResNet} by excluding batch normalization layers, ReLU layers and loss layer. If taking all these layers into consideration, the total number of layers will be $1 + 1 + 1 + 1 + 8 \times (3+4+23+3)  + 1 + 1 + 1 + 1 = 272$.

\newcommand{\blockb}[3]{\multirow{3}{*}{\(\left[\begin{array}{c}\text{1$\times$1, #2}\\[-.1em] \text{3$\times$3, #2}\\[-.1em] \text{1$\times$1, #1}\end{array}\right]\)$\times$#3}
}
\renewcommand\arraystretch{1.1}
\setlength{\tabcolsep}{3pt}
\begin{table}[h]
\caption{CNN architecture for ILSVRC2015. }
\begin{center}
\resizebox{0.7\linewidth}{!}{
\begin{tabular}{c|c|c}
\hline
layer name & output size & 103-layer \\
\hline
conv1 & 112$\times$112 & 7$\times$7, 64, stride 2, padding 3\\
\hline
pool1 & 56$\times$56 & 3$\times$3, stride 2, padding 1\\
\hline
\multirow{3}{*}{ResNet1} &  \multirow{3}{*}{56$\times$56}  & \blockb{256}{64}{3}  \\ 
&  & \\
&  & \\
\hline
\multirow{3}{*}{ResNet2} &  \multirow{3}{*}{28$\times$28}  & \blockb{512}{128}{4}  \\ 
&  & \\
&  & \\
\hline
\multirow{3}{*}{ResNet3} &  \multirow{3}{*}{14$\times$14}  & \blockb{1024}{256}{23}  \\ 
&  & \\
&  & \\
\hline
\multirow{3}{*}{ResNet4} &  \multirow{3}{*}{7$\times$7}  & \blockb{2048}{512}{3}  \\ 
&  & \\
&  & \\
\hline
pool2 & 1$\times$1  & average pool 7$\times$7  \\
\hline
fc & 1000 & fc size 2048$\times$1000  \\
\hline
\end{tabular}
}
\end{center}
\vspace{-.5em}
\label{tab:mililsvrc}
\vspace{-.5em}
\end{table}

During training of this MILCNN, we notice that multiple instance learning loss function is slower than softmax with entropy loss function. So instead of training MILCNN from scratch, we use Softmax with entropy loss function for pretraining. In details, we set a weight decay as 0.0001 and momentum as 0.9. The initial learning rate is 0.1 and reduces to 0.01, 0.001 at 30 epochs and 60 epochs. We choose same weights initialization method in~\cite{He2015}.
The best model has achieved 21.08\% top1 error rate with the aforementioned configurations. Then we switch to our designed multiple instance learning loss function and $\lambda$ is set 0.001. 
We form image bags by sampling 5 $224 \times 224$ regions from images of resolution $256 \times 256$. 
Compared to the original deep residual networks~\cite{ResNet}, where the best performance is achieved at 152-layer deep network, our structure only contains 103 layers due to the limitation of GPU memories (It will take about 11G memory for batch size 64).

\begin{table}[h]
  \caption{Test set error rates for ILSVRC2015 of various methods}
  \label{tab:ilsvrc2015}
  \vskip 0.15in
  \begin{center}
  \begin{small}
  \begin{sc}
  \begin{tabular}{p{4.5cm} c | c}
    \hline
    Method            & Top-1 Error & Top-5 Error \\
    \hline
    VGG~\cite{VGG} (ILSVRC'14) & - & 8.43\\
    GoogLeNet~\cite{GoogleNet} (ILSVRC'14) & - & 7.89\\
    \hline
    VGG~\cite{VGG} \footnotesize (v5) & 24.4 & 7.1\\
    PReLU-net~\cite{He2015} & 21.59 & 5.71 \\
    BN-inception \cite{BN} & 21.99 & 5.81 \\
    ResNet(101-layer)~\cite{ResNet} & 19.87 & 4.60 \\       
    ResNet(152-layer)~\cite{ResNet} & 19.38 & 4.49 \\
    \hline
    \hline
    ResNet(103-layer)          & 21.08 & 5.35 \\
    MILCNN(103-layer)          & 20.78 & 5.30 \\
    
  \end{tabular}
  \end{sc}
  \end{small}
  \end{center}
  \vskip -0.1in
\end{table}

From Table~\ref{tab:ilsvrc2015}, our implementation of residual networks has achieved the state-of-the-art performance compared to other methods for single model evaluation on validation dataset. Our single MILCNN model is able to further improve the performance to 5.30\% and it is reaching the human-level performance 5.10\%~\cite{imageNet}.   

\begin{figure*}[t]
\centering
\includegraphics[width=6.5in]{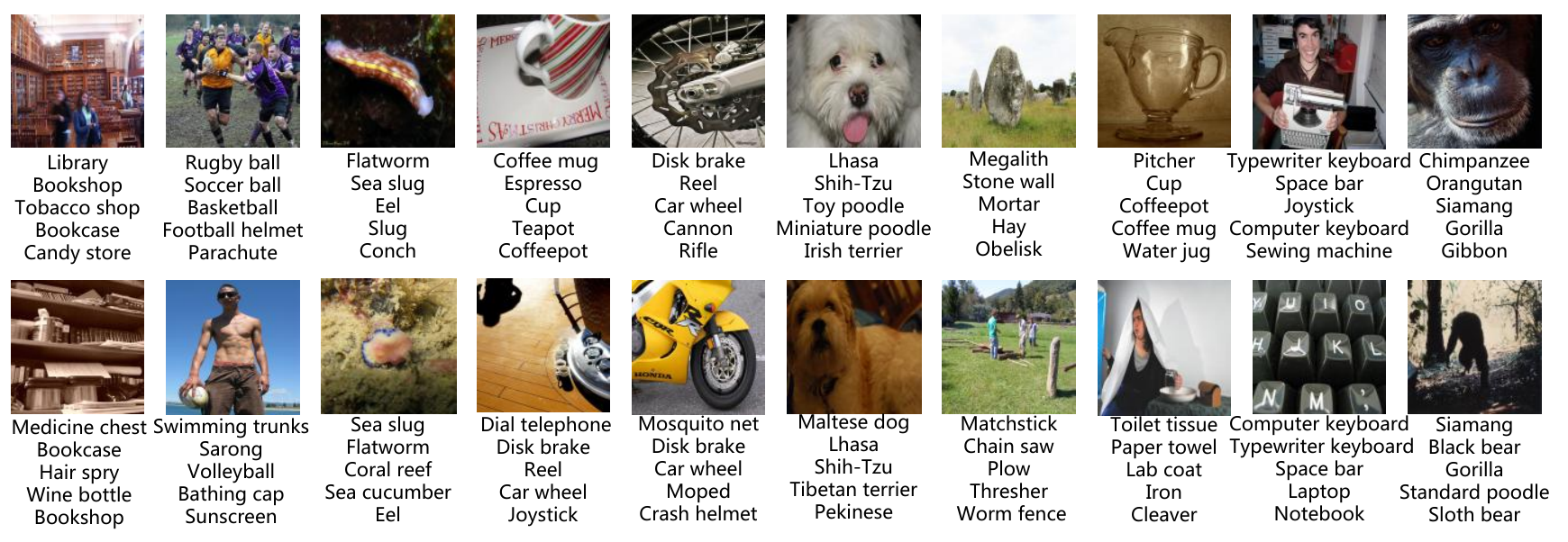}
\caption{{\bf Visualization}: First row are examples validation images successfully classified with top 1 predicted labels. Second row are examples with incorrect top 1 predicted labels. Each column shares same ground-truth label. }
\label{fig:vis}
\end{figure*}

Fig~\ref{fig:vis} shows some predictions of validation images. We randomly select some categories and each category has two images: top one is successfully classified with top 1 predicted labels, bottom one is incorrectly classified with top 1 predicted labels. Since MILCNN still choose relatively large regions to form bags, our algorithm still makes mistakes in cases that have relatively small objects. 
For example, both images in the eighth column are in ``Pitcher" category, but the bottom ``Pitcher" is too small to be visible.
MILCNN can also make less mistakes when more context information is considered, such as the images in the seventh row. The bottom image is labeled as ``Matchstick" while the ground truth is ``Megalith". If taking the context information(positions and scale of person in the image), MILCNN is supposed to rule out the possibility of ``Matchstick" at least.  


\section{Conclusion}

We provided a weakly supervised framework for image recognition by combining multiple instance learning loss and deep residual networks. 
We presented mathematical formulation for how to incorporate the concept of multiple instance learning to a deep learning architecture. Besides, we showed state-of-the-art performance on both low-resolution CIFAR datasets and high-resolution ILSVRC2015 classification dataset.  

\section*{Acknowledgment}
The authors would like to thank Sony Electronics Inc. for their generous funding of this work.  
\IEEEtriggeratref{13}

\bibliographystyle{IEEEtran}
\bibliography{root}

\end{document}